# Intelligent Autonomous Things on the Battlefield

**Alexander Kott**

U.S. Army Research Laboratory, Adelphi, MD, USA
alexander.kott1.civ@mail.mil

**Ethan Stump**

U.S. Army Research Laboratory, Adelphi, MD, USA
ethan.a.stump2.civ@mail.mil

**Abstract**

Numerous, artificially intelligent, networked things will populate the battlefield of the future, operating in close collaboration with human warfighters, and fighting as teams in highly adversarial environments. This chapter explores the characteristics, capabilities and intelligence required of such a network of intelligent things and humans – Internet of Battle Things (IOBT). The IOBT will experience unique challenges that are not yet well addressed by the current generation of AI and machine learning.

## 3.1 Introduction

Internet of Battle Things is the emerging reality of warfare. A variety of networked intelligent systems—"things"—will continue to proliferate on the battlefield where they will operate with varying degrees of autonomy. Intelligent things will not be a rarity but a ubiquitous presence on the future battlefield (Scharre 2014).

Most of such intelligent things will not be too dissimilar from the systems we see on today's battlefield, such as unattended ground sensors, guided missiles (especially the fire-and-forget variety) and, of course, unmanned aerial systems (UAVs). They will likely include physical robots ranging from a very small size (such as insect-scale mobile sensors) to large vehicles that can carry troops and supplies. Some will fly; others will crawl, walk or ride. Their functions will be diverse. Sensing (seeing, listening, etc.) the battlefield will be one common function. Numerous small, autonomous sensors can cover the battlefield and provide an overall awareness to warfighters that is reasonably complete and persistent (Fig. 3.1).

Other things might act as defensive devices; e.g., autonomous active protection systems (Freedberg 2016). Finally, there will be munitions that are intended to impose physical or cyber effects on the enemy. These will not be autonomous; instead, they will be controlled by human warfighters. This assumes that the combatants of that future battlefield will comply with a ban on offensive autonomous weapons beyond meaningful human control. Although the US Department of Defense already imposes strong restrictions on autonomous and semi-autonomous weapon systems (Hall 2017), nobody can predict what other countries might decide on this matter.

In addition to physical intelligent things, the battlefield—or at least the cyber domain of the battlefield—will be populated with disembodied cyber robots. These will reside within various computers and networks, and will move and act in cyberspace. Just like physical robots, cyber robots will be employed in a wide range of roles. Some will protect communications and information (Stytz et al. 2005) or will fact-check, filter, and fuse information for cyber situational awareness (Kott et al. 2014). Others will defend electronic devices from the effects of electronic warfare using actions such as the creation of informational or electromagnetic deceptions or camouflage. Yet others will act as situation analysts and decision advisers to humans or physical robots. In addition to these defensive or advisory roles, cyber robots might also take on more assertive functions, such as executing cyber actions against enemy systems (Fig. 3.2).

In order to be effective in performing these functions, battle things will have to collaborate with each other, and also with human warfighters. This collaboration will require a significant degree of autonomous self-organization and acceptance of a variety of relations between things and humans (e.g., from complete autonomy of an unattended ground sensor to tight control of certain systems), and these modes will have to change as needed. Priorities, objectives and rules of engagement will change rapidly, and intelligent things will have to adjust accordingly (Kott et al. 2016).

Clearly, these requirements imply a high degree of intelligence on the part of the things. Particularly important is the necessity to operate in a highly adversarial environment; i.e., an intentionally hostile and not merely randomly dangerous world. The intelligent things will have to constantly think about an intelligent adversary that strategizes to deceive and defeat them. Without this adversarial intelligence, the battle things will not survive long enough to be useful.

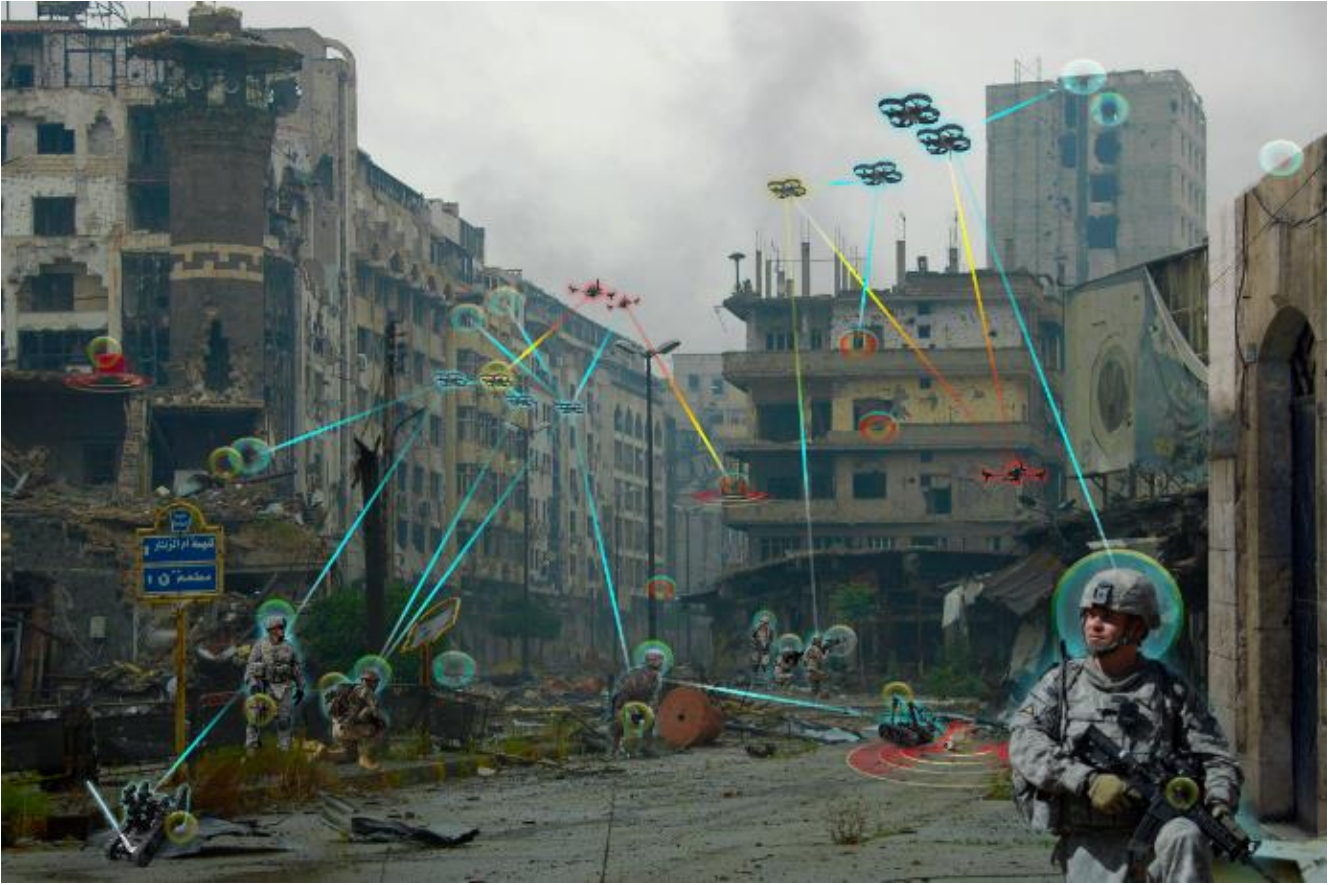

*Figure 3.1 Networked teams of intelligent things and humans will operate in an extremely complex, challenging environment that is unstructured, unstable, rapidly changing, chaotic, rubble-filled, adversarial, and deceptive.*

## 3.2 The Challenges of Autonomous Intelligence on the Battlefield

The vision—or rather, the emerging reality—of the battlefield populated by intelligent things portends a multitude of profound challenges. The use of AI for battlefield tasks has been explored on multiple occasions (e.g., Rasch et al. 2002), and though it makes things individually and collectively more intelligent, it also makes the battlefield harder to understand and manage. Human warfighters have to face a much more complex, more unpredictable world where things have a mind of their own and perform actions that may appear inexplicable to humans. Direct control of such intelligent things becomes impossible or limited to cases of decisions about whether to take a specific destructive action.

On the other hand, humans complicate the life for intelligent things. Humans and things think differently. Intelligent things, in the foreseeable future, will be challenged in understanding and anticipating human intent, goals, lines of reasoning and decisions. Humans and things will remain largely opaque to each other and yet, things will be expected to perceive, reason and act while taking into account the social, cognitive and physical needs of their human teammates. Furthermore, things will often deal with humans who are experiencing extreme physical and cognitive stress, and may therefore behave differently from what can be assumed from observing humans under more benign conditions.

An intelligent thing will need to deal with a world of astonishing complexity. The sheer number and diversity of things and humans within the IOBT will be enormous. For example, the number of connected things within a future Army brigade is likely to be several orders of magnitude greater than in current practice. This number, however, is just the beginning. Consider that intelligent things belonging to such a brigade will inevitably interact—willingly or unwillingly—with things owned and operated by other parties, such as those of the adversary or owned by the surrounding civilian population. If the brigade operates in a large city where each apartment building can contains thousands of things, the overall universe of connected things grows dramatically. Millions of things per square kilometer is not an unreasonable expectation (Fig. 3.2).

The above scenario also points to a great diversity of things within the overall environment of the battlefield. Things will come from different manufacturers, with different designs, capabilities and purposes, configured or machine-learned differently, etc. No individual thing will be able to use pre-conceived (pre-programmed, pre-learned, etc.) assumptions about behaviors or performance of other things it meets on the battlefield. Instead, behaviors and characteristics will have to be learned and updated autonomously and dynamically during the operations. This includes humans (yes, humans are a species of things, in a way) and therefore the behaviors and intents of humans, such as friendly warfighters, adversaries, civilians and so on will have to be continually learned and inferred.

The cognitive processes of both things and humans will be severely challenged in this environment of voluminous and heterogeneous information. Rather than the communications bandwidth, the cognitive bandwidth may become the most severe constraint. Both humans and things seek information that is well-formed, reasonably sized, essential in nature, and highly relevant to their current situation and mission. Unless information is useful, it does more harm than good. The trustworthiness of the information and the value of information arriving from different sources (especially other things) will be highly variable and generally uncertain. For any given intelligent thing, the incoming information could contain mistakes, erroneous observations or conclusions made by other things, or intentional distortions (deceptive information) produced by an adversary malware residing on friendly things or otherwise inserted into the environment. Both humans and things are susceptible to deception, and humans are likely to experience cognitive challenges when surrounded by opaque things that might be providing them with untrustworthy information (Kott and Alberts 2017).

This situation reminds us that the adversarial nature of the battlefield environment is a concern of exceptional importance, above all others. The intelligent things will have to deal with an intelligent, capable adversary. The adversary will bring about physical destruction, either by means such as gunfire (also known as "kinetic" effects) or by using directed energy weapons. The adversary will jam the communication channels between things, and between things and humans. The adversary will deceive things by presenting them with misleading information. Recent research in adversarial learning comes to mind in this connection (Papernot et al. 2016). Perhaps, most dangerously, the adversary will attack intelligent things by depositing malware on them.

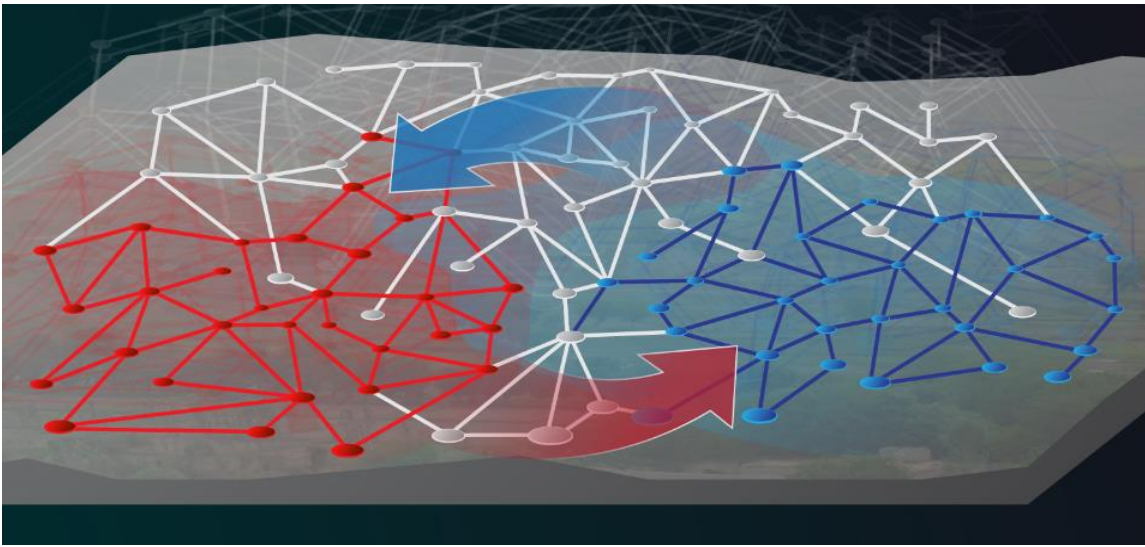

*Figure 1.2 Networks of opponents will fight each other with cyber and electromagnetic attacks of great diversity and volume; most such offensive and defensive actions will be performed by autonomous cyber agents.*

## 3.3 AI will Fight the Cyber Adversary

A key assumption that must be made regarding the IoBT is that in a conflict with a technically sophisticated adversary, IoBT will be a heavily contested battlefield (Kott et al. 2015). Enemy software cyber agents, or malware, will infiltrate our networks and attack our intelligent things. To fight them, things will need artificial cyber hunters - intelligent, autonomous, mobile agents specialized in active cyber defense and residing on IoBT.

Such agents will stealthily patrol the networks, detect the enemy malware while remaining concealed, and then destroy or degrade the enemy malware. They will do so mostly autonomously, because human cyber experts will be always scarce on the battlefield. They will be adaptive because the enemy malware is constantly evolving. They will be stealthy because the enemy malware will try to find and kill them. At this time, such capabilities do not exist but are a topic of research (Theron et al. 2018). We will now explore the desired characteristics of an intelligent autonomous agent operating in the context of IoBT.

Under consideration is a thing—a simple senor or a complex military vehicle—on which one or more computers resides. Each computer contributes considerably to the operation of the thing or systems installed on the thing. One or more of the computers is assumed to have been compromised, where the compromise is either established as a fact or is suspected.

Due to the contested nature of the communications environment (e.g., the enemy is jamming the communications or radio silence is required to avoid detection by the enemy), communications between the thing and other elements of the friendly force can be limited and intermittent. Under some conditions, communications are entirely impossible.

Given the constraints on communications, conventional centralized cyber defense is infeasible. (Here, centralized cyber defense refers to an architecture where local sensors send cyber-relevant information to a central location, where highly capable cyber defense systems and human analysts detect the presence of malware and initiate corrective actions remotely.) It is also unrealistic to expect that the human warfighters in the vicinity of the thing (if they exist) have the necessary skills or time available to perform cyber defense functions for that thing.

Therefore, cyber defense of the thing and its computing devices must be performed by an intelligent, autonomous software agent. The agent (or multiple agents per thing) would stealthily patrol the networks, detect the enemy agents while remaining concealed, and then destroy or degrade the enemy malware. The agent must do so mostly autonomously, without the support or guidance of a human expert.

In order to fight the enemy malware deceptively deployed on the friendly thing, the agent often has to take destructive actions, such as deleting or quarantining certain software. Such destructive actions are carefully controlled by the appropriate rules of engagement and are allowed only on the computer where the agent resides. The agent may also be the primary mechanism responsible for defensive cyber maneuvering (e.g., a mobbing target defense), deception (e.g., redirection of malware to honeypots [De Gaspari et al. 2016]), self-healing (e.g., Azim et al. 2014), and other such autonomous or semi-autonomous behaviors (Jajodia et al. 2011).

In general, the actions of the agent cannot be guaranteed to preserve the integrity of the functions and data of friendly computers. There is a risk that an action of the agent may "break" the friendly computer, disable important friendly software, or corrupt or delete important data. In a military environment, this risk must be balanced against the death or destruction caused by the enemy if an agent's recommended action is not taken.

Provisions are made to enable a remote or local human controller to fully observe, direct and modify the actions of the agent. However, it is recognized that human control is often impossible, especially because of intermittent communications. The agent is therefore able to plan, analyze and perform most or all of its actions autonomously. Similarly, provisions are made for the agent to collaborate with other agents (residing on other computers); however, in most cases, because the communications are impaired or observed by the enemy, the agent operates alone.

The enemy malware and its associated capabilities and techniques evolves rapidly, as does the environment in general, together with the mission and constraints to which the thing is subject. Therefore, the agent is capable of autonomous learning.

Because the enemy malware knows that the agent exists and is likely to be present on the computer, the enemy malware seeks to find and destroy the agent. Therefore, the agent possesses techniques and mechanisms for maintaining a degree of stealth, camouflage and concealment. More generally, the agent takes measures that reduce the probability of its detection by the enemy malware. The agent is mindful of the need to exercise self-preservation and self-defense.

It is assumed here that the agent resides on the computer where it was originally installed by a human controller or by an authorized process. We envision the possibility that an agent may move itself (or a replica of itself) to another computer. However, it is assumed that such propagation occurs only under exceptional and well-specified conditions and only within a friendly network—from one friendly computer to another friendly computer. This situation brings to mind the controversy about "good viruses." Such viruses have been proposed, criticized and dismissed earlier (Muttik 2016). These criticisms do not apply here. This agent is not a virus because it only propagates under explicit conditions within authorized and cooperative nodes. Also, it is used only in military environments, where most of the usual concerns are irrelevant.

# 3.4 AI will Perceive the Complex World

Agents will have to become useful team-mates—not tools—of human warfighters on a highly complex and dynamic battlefield. Figure 3.1 depicts an environment wherein a highly-dispersed team of human and intelligent agents (including but not limited to physical robots) is attempting to access a multitude of highly heterogeneous and uncertain information sources and use them to form situational awareness and make decisions (Kott et al. 2011), while simultaneously trying to survive extreme physical and cyber threats. They must be effective in this unstructured, unstable, rapidly changing, chaotic and rubble-filled adversarial environment; learning in real-time under extreme time constraints, with only a few observations that are potentially erroneous, with uncertain accuracy and meaning, or are even intentionally misleading and deceptive (Fig. 3.3).

It is clearly far beyond the current state of AI to operate intelligently in such an environment and with such demands. In particular, Machine Learning, an area that has seen dramatic progress in the last decade, must experience major advances in order to become relevant to the real battlefield. Let's review some of the required advances.

Learning with a very small number of samples is clearly a necessity in an environment where friends and enemies continuously change their tactics and the environment itself is highly fluid, rich with details, dynamic and changing rapidly. Furthermore, very few (if any) of the available samples will be labelled, or if so, not in a very helpful manner. This learning stands in stark contrast to the highly influential ImageNet dataset (Deng et. al. 2009) that led to the advent of modern Deep Learning by providing a rich, labeled dataset for a massive 1000-class task. Whereas ImageNet data is labeled and provides one main class per image, real field data is fraught with partially occluded objects and ambiguous detections.

Dealing with limited samples means moving beyond the current state-of-the-art in Deep Learning that seeks to learn efficient representations for entire domains by only allowing each sample to influence the model a very small amount. Embracing classical non-parametric techniques that treat each piece of data as representative of its own local domain is one potential path to sample-efficient learning. Overcoming the never-ending growth of data required to define our model is an area of active research, but by allowing the system to trade accuracy for efficiency, it is possible to keep the cost of learning in check (Koppel et. al. 2017).

Truly intelligent agents, however, will not need to memorize data to make sense of the changing world; rather, they will be able to investigate only a few examples in order to quickly learn how the current environment relates to their experience. This technique of *domain adaptation* (Patel et. al. 2015) promises to allow models trained over exhaustive datasets of benign environments to remain useful on a dynamic battlefield. Whether by updating only a small portion of the model (Chu et.al. 2016), appealing to the underlying geometry of the data manifolds (Fernando et. al. 2013; Gong et. al. 2012), or turning the learning algorithms upon themselves to be trained how to adapt (Long et. al. 2015), future agents will maintain flexible representations of their learned knowledge, amenable to reinterpretation and reuse.

Flexibility of learning and knowledge is crucial: it will undoubtedly be undesirable for an agent to enter a new environment with pre-trained (preconceived), absolute notions of how it should perceive and act. Instead, an agent will always be formulating and solving learning problems; and the role of training is to teach the system how it can do this as efficiently as possible, perhaps even in one learning step (Finn et. al. 2017). This fascinating concept of *meta-learning*, or learning how to learn (Andrychowicz et. al. 2016), allows the agent to finally take advantage of both its knowledge and experience to perceive and interact with the dynamic world and an evolving team and mission.

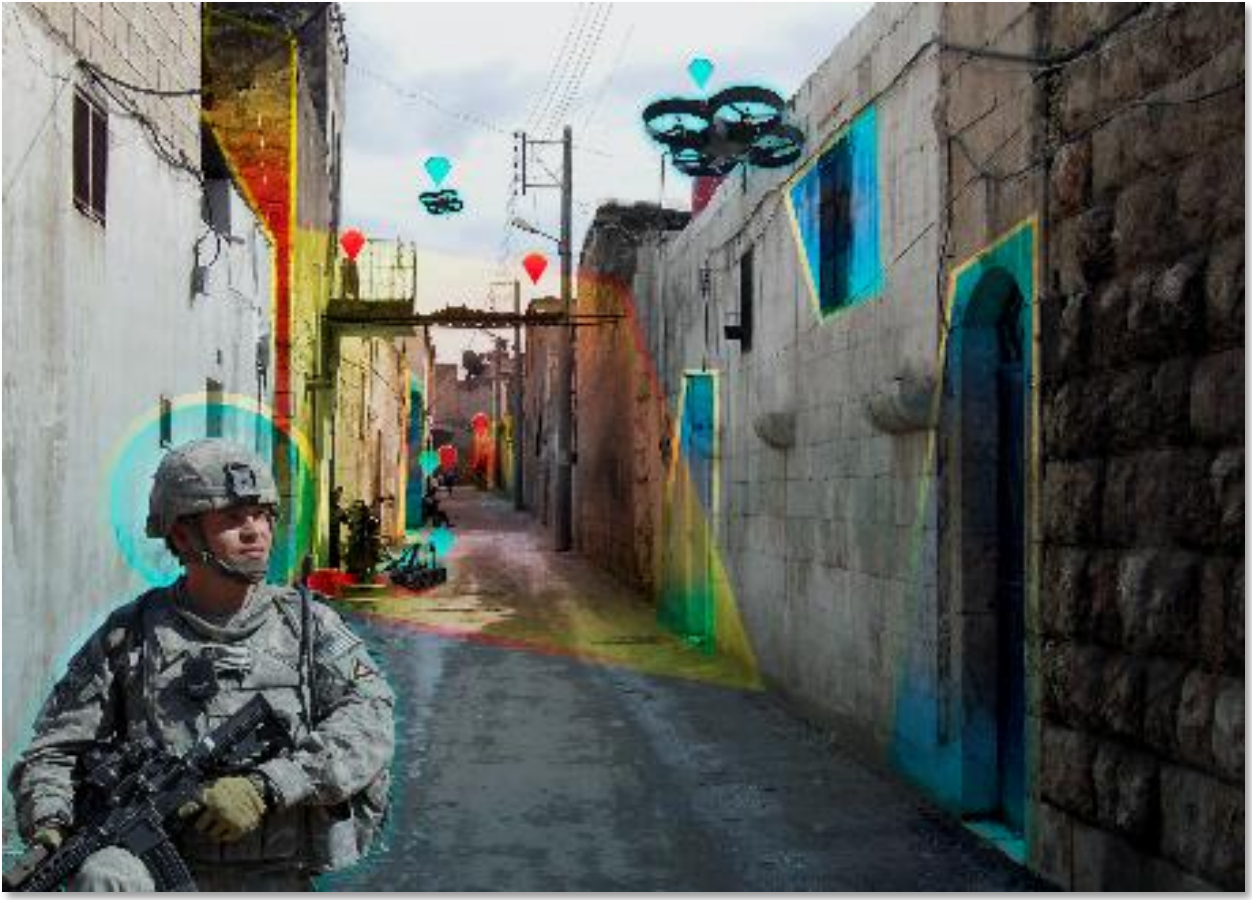

*Figure 3.3 AI-enabled agents—members of a human-agent team—will rapidly learn in ever-changing, complex environments, providing the team's commander with real-time estimates of enemy, reasoning on possible courses of action, and tactically sensible decision.*

The domains that agents must learn and understand are vast and complex. A typical example might be a video snippet of events and physical surroundings for a robot, where the overwhelming majority of elements (e.g., pieces of rubble) are hardly relevant and potentially misleading for the purposes of learning. The information in the samples is likely to be highly heterogeneous in nature. Depending on circumstances, samples might consist of one or more of the following data: still images in various parts of the spectrum (IR, visible, etc.), video, audio, telemetry, solid models of the environment, records of communications between agents, and so on. Multiple modalities offers the potential to allow one type of data to help sort through ambiguities and un-observability in another modality. Moreover, complex and widespread events such as social movements can only truly be understood by aligning and composing these heterogeneous sources to detect underlying patterns (Giridhar et. al. 2017; Gui et. al. 2017).

The danger is that some samples may be misleading in general, even if unintentionally (e.g., an action succeeds even though an unsuitable action is applied) and the machine learning algorithms will have to make the distinction between relevant and irrelevant, instructive and misleading. In addition, some of the samples might be a product of intentional deception by the enemy. In general, issues of Adversarial Learning and Adversarial Reasoning are of great importance (Papernot et al. 2016). Although initial demonstrations that showed how to perturb an image to fool a trained model were seen as an amusement, later demonstrations have shown that not only can these attacks be used in the real environment (Kurakan et. al. 2016), but they do not require insider knowledge of the learning system (Papernot et. al. 2017) and are exceedingly difficult to detect in isolation (Carlini and Wagner, 2017). Robustness to these adversarial actions will come about as the product of engineering principles that treat learning as part of a larger system of models and heterogeneous data that provide avenues to check and attest to the veracity of our models.

## 3.5 AI Enables Embodied Agents

Some intelligent things will be embodied so that they can actively explore and interact with the world, not merely constructs that protect our virtual environment or sensors that vigilantly observe and interpret. AI enables these interactions, allowing things to develop intuition about the laws of physics and to learn how to act optimally to accomplish their missions. In some cases, agents will take our traditional hand-crafted control paradigms and expand them through learning and artificial evolution; in other cases, agents will learn behaviors as a child does, developing motor skills through careful play. The agents will produce dramatic new behaviors that are perfectly suited for their tasks yet display the same adaptability that we find in their perception systems. These creative solutions to embodiment will accomplish another important goal: to find the inflection points in the design and control space where complex physical interactions of body and environment lead to mechanical advantage and efficient cycles that let systems overcome their fundamental limitations on weight and power.

Humans have developed many control schemes that allow us to bend natural processes to our will, and machines must continue these developments, tapping into a capacity for learning and adaptation to magnify these traditional techniques. In some cases, this means identifying pieces of our planning and control systems that are particularly hard to model and letting the system learn them (Richter et. al. 2015). In other cases, we can take a more holistic view of learning to plan and control: recent advances in planning have allowed us to sidestep the challenge of thinking through all of the compounding choices by substituting enumeration with sampling and relying on the mechanisms of probability to give us a path to convergence and optimality (Karaman and Frazzoli, 2011; Tedrake et. al. 2010). Intelligent systems will learn to guide their sampling over time as a way to realize the knowledge they have built about the world and their capabilities while still retaining the ability to be flexible in case their model is not quite right. Modeling probability distributions can be challenging, but learning to transform a simple distribution can accomplish the same goal and provide a path to learnable sampling-based control (Lenz et. al. 2015).

These planning and control approaches are founded on the ability to make predictions about the effect of our actions on the world, but, in a battlefield environment where conditions can deteriorate rapidly and intelligent things may need to continuously adapt to damage, forming these predictions is another task that AI must step in to perform. We can take inspiration from studies of how the human mind develops and learns in order to understand how agents can develop intuition about the physical world (Tenenbaum et. al. 2011); this intuition can be holistic or focused on understanding specific aspects of the world such as rigid body motion (Byravan and Fox, 2017) or how fluids behave (Schenck and Fox, 2016). Through observations and careful experimentation (Pinto et. al. 2016), agents can treat physics as another application of perception algorithms that can take cause and effect data and learn to generalize (Hefny et. al. 2015). These models then become an intimate part of the planning and control process (Boots et. al. 2011; Williams et. al. 2017).

Traditional scientific and design methodology emphasizes the need to model and analyze the parts of a system in isolation, carefully controlling the variables so that the most concise and elegant description of a phenomenon or capability can be developed. In the future where components are the result of learning processes that can radically reshape their functionality, these traditional processes must also evolve. This evolution is already taking place with the advent of so-called *end-to-end learning* approaches that build and simultaneously train the entire processing and decision-making pipeline from perception to action (Levine et. al. 2016; Bojarski et. al. 2016). Simultaneously learning on many tasks at once keeps this process from overspecializing (Devin et. al. 2017) – another example of *meta-learning*. Eventually these techniques will be able to recapture the crucial feature of traditional systems engineering: being able to formally prove correctness (Aswani et. al. 2013).

Despite the future advances in embodied intelligence just discussed, the challenge of limited electrical power will remain a driving factor in deployment for the battlefield. Most successful AI relies on vast computing and electrical power resources including cloud-computing reach-back when necessary. The battlefield AI, on the other hand, must operate within the constraints of edge devices, such as small sensors, micro-robots, and the handheld radios of warfighters. This situation means that computer processors must be relatively light and small, and as frugal as possible in the use of electrical power. One might suggest that a way to overcome such limitations on computing resources available directly on the battlefield is to offload the computations via wireless communications to a powerful computing resource located outside of the battlefield (Kumar and Lu, 2010). Unfortunately, it is not a viable solution because the enemy's inevitable interference with friendly networks will limit the opportunities for use of reach-back computational resources. We must turn to techniques to trade precision for power to keep computing at the edge (Gupta et. al. 2011; Iandola et. al. 2016) or encode mature AI procedures directly into circuitry (Zhang et. al. 2017). Perhaps the systems will again lead the way, just as they did when they used meta-learning to learn adaptability by obtaining the correct balance between power and accuracy and tuning their own algorithms to realize it.

## 3.6 Coordination Requires AI

It is essential that agents work collectively to sense and explore the battlefield: the size and scope of the challenge demand it and agents can fall prey to hazards or adversaries at any time. Humans are masters of coordination, able to work together to accomplish large tasks using an array of different modes of communication, from explicit instructions to codewords to an unspoken understanding of team roles. Coordination on the battlefield will require agents to use all of these mechanisms and more, taking inspiration from nature around us and also from applied logic to demonstrate the ability to share and sequence. Underlying all of this coordination will be the collective activity necessary to enable the communications networks upon which higher-level coordination will depend.

A certain amount of abstraction is required as teams become large and tracking and managing every agent's identity through changing and merging battlefield roles becomes impossible. These abstractions are exemplified by the complex emergent behavior of swarms of fish and birds that move with a purpose but have no explicit guidance. It has been shown that these phenomena can arise from the interactions of simple rules between neighboring agents (Jadbabaie et. al. 2003) and that they are robust, able to maintain cohesion even as neighbors come and go (Olfati-Saber et. al. 2004; Ren and Beard 2005; Tanner et. al. 2007). Not only can these swarms move together, they can also explore and manipulate the world (Berman et. al. 2011). These insights are grounded in application of graph theory to dynamic systems (Mesbahi and Egerstedt 2010); all of this work on understanding and replicating swarms showed how we can synthesize and understand global properties by studying local ones (De Silva et. al. 2005).

It is not enough to just observe the emergent behavior; we must actively control the emergent behavior to realize the vision of coordination that can scale and adapt to meet battlefield challenges. The challenge lies in the fact that, at a suitable level of abstraction, agents have no identity, yet we must allocate and control them toward complex tasks. Again, the answers lie in stochasticity: allowing each agent to randomly determine its own actions but with probabilities proportional to the number of agents required breaks the need for identity by letting each agent self-determine whether it will help (Berman et. al. 2009). These probabilities can be further adapted in a closed-loop fashion to shape the distributions (Mather and Hsieh, 2011).

The final step in controlling swarms is to embrace the heterogeneity of platforms and sensors that we can employ and let agents specialize to their suited tasks while still retaining a degree of anonymity (Kott and Abdelzaher 2014).

A team that includes multiple warfighters and multiple artificial agents must be capable of distributed learning and reasoning. Besides distributed learning, these include such challenges as: multiple decentralized mission-level task allocations; self-organization, adaptation, and collaboration; space management operations; and joint sensing and perception. Commercial efforts to date have been largely limited to single platforms in benign settings. Military-focused programs like Micro Autonomous Systems and Technology (MAST) Collaborative Technology Alliance (CTA) (Piekarski et al. 2017) have been developing collaborative behaviors for UAVs. Ground vehicle collaboration is challenging and is largely still at the basic research level at present. In particular, to address such challenges, a new collaborative research alliance called Distributed and Collaborative Intelligent Systems and Technology (DCIST) has been initiated (https://dcist-cra.org/). Note that the battlefield environment imposes yet another complication: because the enemy interferes with communications, all of this collaborative, distributed AI must work well, even with limited, intermittent connectivity.

## 3.7 Humans in the Ocean of Things

In this vision of the future warfare, a key challenge is to enable autonomous systems and intelligent agents to effectively and naturally interact across a broad range of warfighting functions. Human-agent collaboration is an active ongoing research areathat must address such issues as trust and transparency, common understanding of shared perceptions, and human-agent dialog and collaboration (Kott and Alberts 2017).

One seemingly relevant technology is Question Answering—the system's ability to respond with relevant, correct information to a clearly stated question. Successes of question-answering commercial technologies are indisputable. They work well for very large, stable, and fairly accurate volumes of data (e.g., encyclopedias). But such tools do not work for rapidly changing battlefield data, which is also distorted by an adversary's concealment and deception. They cannot support continuous, meaningful dialog in which both warfighters and artificially intelligent agents develop a shared situational awareness and intent understanding. Research is being performed to develop human-robotic dialog technology for warfighting tasks using natural voice, which is critical for reliable battlefield teaming.

A possible approach to developing the necessary capabilities—both human and AI—is to train a human-agent team in immersive artificial environments. This training requires building realistic, intelligent entities in immersive simulations. Training (for humans) and learning (for agents) experiences must exhibit a high degree of realism to match operational demands. Immersive simulations for human training and machine learning must have physical and sociocultural interactions with high fidelity and a realistic complexity of the operational environment. These include realistic behaviors of human actors (friendly warfighters, enemies, non-combatants), and interactions and teaming with robots and other intelligent agents. In today's video games, these interactions are limited and not suitable for simulating real battlefield. Advances in AI are needed to drive the character behaviors that are truly realistic, diverse, and intelligent.

To this end, some of the cutting-edge efforts in computer-generation of realistic virtual characters are moving toward what would be needed to enable realistic interactions in an artificial immersive battlefield. For example, Hollywood studios sought out the Army-sponsored Institute for Creative Technologies (http://ict.usc.edu/) on multiple occasions to create realistic avatars of actors. These technologies enable film creators to digitally insert an actor into scenes, even if that actor is unavailable, much older or younger, or deceased. This is how the actor Paul Walker was able to appear in "Fast and Furious 7," even though he died partway into filming (CBS News 2017).

## 3.8 Summary

Intelligent things—networked and teamed with human warfighters—will be a ubiquitous presence on future battlefields. Their appearances, roles and functions will be highly diverse. The artificial intelligence required for such things will have to be significantly greater than that provided by today's AI and machine learning technologies. The adversarial—strategically and not randomly dangerous—nature of the battlefield is a key driver of these requirements. Complexity of the battlefield, including the complexity of collaboration with humans, is another major driver. Cyber warfare will assume a far greater importance, and AI will have to fight cyber adversaries. Major advances in areas such as adversarial learning and adversarial reasoning will be required. Simulated immersive environments may help to train the humans and AI.

## Disclaimers